\documentclass[conference]{IEEEtran}
\pdfoutput=1
\makeatletter
\def\ps@headings{%
\def\@oddhead{\mbox{}\scriptsize\rightmark \hfil \thepage}%
\def\@evenhead{\scriptsize\thepage \hfil \leftmark\mbox{}}%
\def\@oddfoot{}%
\def\@evenfoot{}}
\makeatother
\usepackage{amsmath}
\usepackage{amssymb}
\usepackage{bm}
\usepackage{multirow}
\usepackage[bottom=1.05in,top=0.755in, right=0.635in, left=0.65in]{geometry}

\usepackage{mathtools,xcolor}
\usepackage{todonotes}
\newcounter{todocounter}

\usepackage{amsthm}
\usepackage{multicol}
\usepackage{cite}
\usepackage{float}

\restylefloat{table}

\usepackage{algorithm}
\usepackage{algcompatible}
\usepackage{soul}

\usepackage{caption}

\ifCLASSOPTIONcompsoc
\usepackage[caption=false,font=normalsize,labelfont=sf,textfont=sf]{subfig}
\else
\usepackage[caption=false,font=footnotesize]{subfig}
\fi
 
\newtheorem{theorem}{Theorem}

\newtheorem{definition}{Definition}

\makeatletter
\def\ALG@special@indent{%
    \ifdim\ALG@thistlm=0pt\relax
        \hskip-\leftmargin
    \else
        \hskip\ALG@thistlm
    \fi
}

\newcommand{\Serverexe}[1]{\item[]\noindent\ALG@special@indent  \textbf{Server executes:}\ #1}

\newcommand{\VRDPClientUpdate}[1]{\item[]\noindent\ALG@special@indent \textbf{VRDP-Update($\mathbf{x}^t$, $\mathbf{m}^t$):}\ #1}
\makeatother

\makeatletter

\newcommand{\returnx}{\item\noindent{return}\ }

\makeatother

\title{Byzantine-Robust Federated Learning with Variance Reduction and Differential Privacy}

\author{
\IEEEauthorblockN{Zikai Zhang, Rui Hu}
\IEEEauthorblockA{Department of Computer Science and Engineering, University of Nevada, Reno, Reno, NV, 89557}
zikaizhang@nevada.unr.edu, ruihu@unr.edu}

\begin{document}

\maketitle

\begin{abstract}
Federated learning (FL) is designed to preserve data privacy during model training, where the data remains on the client side (i.e., IoT devices), and only model updates of clients are shared iteratively for collaborative learning. However, this process is vulnerable to privacy attacks and Byzantine attacks: the local model updates shared throughout the FL network will leak private information about the local training data, and they can also be maliciously crafted by Byzantine attackers to disturb the learning. In this paper, we propose a new FL scheme that guarantees rigorous privacy and simultaneously enhances system robustness against Byzantine attacks. Our approach introduces sparsification- and momentum-driven variance reduction into the client-level differential privacy (DP) mechanism, to defend against Byzantine attackers. The security design does not violate the privacy guarantee of the client-level DP mechanism; hence, our approach achieves the same client-level DP guarantee as the state-of-the-art. We conduct extensive experiments on both IID and non-IID datasets and different tasks and evaluate the performance of our approach against different Byzantine attacks by comparing it with state-of-the-art defense methods. The results of our experiments show the efficacy of our framework and demonstrate its ability to improve system robustness against Byzantine attacks while achieving a strong privacy guarantee.

\end{abstract}


\section{Introduction}
In the Internet of Things (IoT) era, vast volumes of data are continuously generated and collected by myriad devices embedded in our everyday environments. In conventional centralized machine learning (ML) systems, data from various users (or their edge devices) are gathered and processed on a central server for model training, which raises significant concerns about data privacy. 
Federated learning (FL), a novel distributed ML paradigm, presents an appealing solution to protect users' privacy\cite{fl_original}. In FL, edge devices collaboratively learn a shared ML model while keeping their data locally, thereby mitigating privacy leakage and reducing the need for data transmission. The enhanced privacy and efficiency offered by FL have prompted its application in various data-sensitive IoT scenarios~\cite{iotw}, including communication~\cite{hu2022energy}, network~\cite{commz}, business~\cite{hu2020trading}, etc. 

However, vanilla FL has been demonstrated to be susceptible to various security and privacy vulnerabilities~\cite{advl, hu2019targeted}, with \textit{Byzantine attack}~\cite{fang, agr} and \textit{privacy inference attack}~\cite{shokri2017membership} standing out as two of the most prevalent. In Byzantine attacks, a number of devices (or clients) participating in FL can act maliciously and disrupt the training of the shared ML model. In a typical training round of FL, devices download the shared global model maintained by a central server and update it locally using their own data, after which the server collects and aggregates these local model updates to obtain the global model update. This process, however, exposes a vulnerability to malicious devices. Such devices can subtly manipulate their local model updates to prevent or mislead the convergence of the shared global model. Notably, it has been shown that even a single malicious device can alter the trained global model\cite{krum}. 
In the privacy inference attack, both devices and the server involved in FL can infer information about the local data of the devices by observing these model updates, thereby posing a privacy risk. For example, \cite{nasr2019comprehensive} shows that the malicious server and device can infer whether a particular individual data record was included in the training dataset with high accuracy.

Current countermeasures for Byzantine attacks and privacy inference attacks in FL mainly focus on two distinct directions: 1) improving the robustness of the global model against malicious model updates and 2) providing privacy protection for each device's local dataset. To provide Byzantine robustness, existing works emphasize designing Byzantine-robust aggregators~\cite{trmean_median, krum, bulyan} to filter out or moderate suspicious local updates uploaded by Byzantine devices at the central server. For privacy protection, the differential privacy (DP) mechanism \cite{dp_original} is a common practice that adds calibrated noise to the model updates before sharing, achieving a rigorous privacy guarantee for each device in FL. Besides, a few works attempt to orchestrate Byzantine robustness and privacy; however, they typically struggle to excel in both aspects simultaneously or operate under the assumption of a trusted central server. For instance, \cite{can-you-really} demonstrates that incorporating ``weak" differential privacy can mitigate poisoning attacks in distributed systems and enhance model robustness. However, the applied differential privacy noise in their work is deemed ``weak" and does not provide a meaningful privacy guarantee. Flame~\cite{flame} is a state-of-the-art defense method against poisoning attacks. It utilizes clustering and adaptive clipping strategies, combined with client-level differential privacy, to enhance system robustness. However, it falls short of achieving a balance between privacy and robustness. Secure aggregation~\cite{bonawitz2017practical} is co-designed with Byzantine robustness aggregators, but secure aggregation assumes the server is trusted.

In this work, our objective is to design a federated learning system that is both Byzantine-robust and privacy-preserving. {We assume that both the clients and the server are semi-honest.} Specifically, the central server and clients are "honest-but-curious," which means they honestly follow the training protocol but may be curious about the private local datasets; and a fraction of clients (less than 50\%) are also malicious/Byzantine, who do not honestly follow the local model training protocol and can generate arbitrary malicious local updates. Inspired by \cite{flame}, we explore the potential advantages of using the DP mechanism for Byzantine robustness, particularly when a strong privacy guarantee is required. We target the cross-device FL system with client-level DP and aim to improve its robustness against Byzantine attacks without violating its privacy guarantee. The main challenge here is the high variance of the aggregated model update introduced by local stochastic gradients of benign clients, Byzantine attackers, and DP noise.

To address this challenge, we introduce a novel FL scheme that combines sparsification- and momentum-driven variance reduction techniques with the client-level DP mechanism. We first reduce the variance of the local stochastic gradient training process using local momentum, which is proved to be essential for improving the robustness. Then, we integrate top-$k$ model sparsification into the commonly used client-level DP mechanism without violating its privacy guarantee. This integration allows us to reduce the impact of Byzantine and DP perturbations on the aggregated model update. Through extensive experiments on multiple datasets, our method demonstrates remarkable performance in terms of both robustness and privacy. Particularly, our approach surpasses existing Byzantine-robust aggregators, showcasing its superiority. When applied to the Fashion-MNIST dataset with 20\% Byzantine clients, our approach achieves an impressive testing accuracy improvement of up to +67.34\%, while ensuring that the privacy loss of each client is bounded by 1.01 after the training. The main contributions of this work can be summarized as follows:

\begin{itemize}

\item We propose a novel FL approach named FedVRDP to defend against Byzantine model poisoning attacks and privacy inference attacks in FL. FedVRDP is designed to achieve Byzantine robustness and differential privacy simultaneously with minimal modifications to the existing commonly-used FL framework FedAvg\cite{fl_original}. 

\item FedVRDP combines sparsification- and momentum-driven variance reduction techniques with the client-level DP mechanism. We provide a theoretical analysis to illustrate the variance sources present in the aggregated local updates of the DP-based FedAvg baseline when Byzantine clients are involved and mitigate the pivotal factors that contribute to increased variance.


\item In FedVRDP, the design for improving robustness against Byzantine attackers does not compromise the privacy guarantee of the client-level DP mechanism in FL. As a result, FedVRDP attains the same client-level DP assurance as the state-of-the-art (SOTA) FL scheme with client-level DP. Simultaneously, FedVRDP markedly enhances the system's resilience against Byzantine attacks, thereby upholding both robustness and privacy simultaneously. 

\item We evaluate our approach across IID and non-IID datasets through extensive experiments, and compare the results with those of SOTA defense methods. The results underscore our approach's effectiveness in improving Byzantine robustness and privacy protection.
\end{itemize}
The rest of the paper is organized as follows. Backgrounds and preliminaries on FL and DP are described in Section~\ref{sec:pre}. Section~\ref{sec:sys-mod} introduces the problem setting and discusses the existing solutions. Section~\ref{sec:our} presents our proposed method FedVRDP, and Section~\ref{sec:exp} shows the experimental results. Finally, Section~\ref{sec:related} reviews the related work, followed by the conclusion in Section~\ref{sec:con}.

\section{Backgrounds and Preliminaries}\label{sec:pre}
\subsection{FL System \& FedAvg} \label{sec:fedvag}
We consider a typical FL system of $n$ clients (e.g., IoT devices) and a central server, and the clients collaboratively train a global model $\mathbf{x} \in \mathbb{R}^d$ with dimension $d$ under the coordination of the server. Assume each client $i\in [n]$ holds a local private dataset $D_i$. The goal of FL is to solve the following optimization problem:
\begin{align}\label{fed_obj}
\min_{\mathbf{x} \in \mathbb{R}^d} f(\mathbf{x}) := \frac{1}{n}\sum_{i\in[n]} f_{i}(\mathbf{x}). 
\end{align}
where $f_{i}(\mathbf{x}) := \mathbb{E}_{z \in  D_i}[l(\mathbf{x};z)]$ represents the local loss of client $i$, and $l(\mathbf{x};z)$ is the loss as a function of the model parameter $\mathbf{x}$ and a datapoint $z$ sampled from ${D}_i$. Note that, for $i\neq j$, the datasets ${D}_i$ and ${D}_j$ may have different distributions. 

To solve the problem in \eqref{fed_obj}, the whole system runs $T$ rounds of FL training protocol. Initially, the server stores a global model $\mathbf{x}^0$. In the $t$-th round, the server randomly selects a subset $\mathcal{S}^t$ of $s$ clients and sends them the latest global model $\mathbf{x}^t$. The selected clients then update the global model $\mathbf{x}^t$ using their local datasets. For example, in the classic and most widely-used FL algorithm, Federated Averaging ({FedAvg}) \cite{fl_original}, each selected client performs $\tau$ iterations of SGD to update the global model as follows:
\begin{equation}\label{eqn:local-sgd}
\mathbf{x}_i^{t,r+1} = \mathbf{x}_i^{t,r} - \eta \cdot \bm{g}_i^{t,r}, \forall r = 0, \ldots, \tau - 1,
\end{equation}
Here, $\mathbf{x}_i^{t, r}$ represents the local model of client $i$ during local training, which is initialized as $\mathbf{x}_i^{t, 0} = \mathbf{x}^{t}$. $\eta$ is the non-negative local learning rate, and $\bm{g}_i^{t,r}:=(1/B)\sum_{z\in\xi_i^{t,r}}\nabla l(\mathbf{x}_i^{t,r},z)$ is the stochastic gradient over a mini-batch $\xi_i^{t,r}$ of $B$ data points sampled from $D_i$. After local training, the clients compute the model update $\Delta_i^t:=\mathbf{x}^{t}-\mathbf{x}_i^{t,\tau} $ and send it to the server. In the $t$-th round, the server updates the global model $\mathbf{x}^{t}$ as follows:
\begin{equation}\label{global_aggregation}
   \mathbf{x}^{t+1} = \mathbf{x}^{t} - \frac{1}{s} \sum_{i\in\mathcal{S}^t} \Delta_i^t.
\end{equation}
where $s=|\mathcal{S}^t|$ is the number of selected clients at round $t$. 

\subsection{Differential Privacy (DP) \& Client-level DP Mechanism} \label{sec:client-dp}
The classic notion of DP, $(\epsilon, \delta)$-DP, is defined as follows:
\begin{definition}[$(\epsilon,\delta)$-DP\cite{dp_original}]\label{DP} 
Given privacy parameters $\epsilon >0$ and $0\leq \delta < 1$, a randomized mechanism $\mathcal{M}$ satisfies $(\epsilon,\delta)$-DP if for any two adjacent datasets $D, D^{\prime}$ and any subset of outputs ${O} \subseteq \text{range}(\mathcal{M})$,
$
\Pr[\mathcal{M}(D) \in {O}] \leq e^{\epsilon} \Pr[\mathcal{M}(D^{\prime}) \in {O}] + \delta.
$
\end{definition}
Let $D$ denote the union of all local datasets of the clients in FL. For client-level DP, two datasets $D$ and $D^{\prime}$ are adjacent datasets if ${D} \cup \{D_j\}$ or ${D}\setminus\{D_j\}$ is identical to $D^{\prime}$ for a local dataset $D_j$. In other words, $D$ and $D^{\prime}$ differ by at most one local dataset. The client-level DP provides rigorous privacy protection to the entire local dataset of a client in the setting of FL, such that the maximum private information that an arbitrary privacy attacker can infer is bounded.  

\textit{Client-level DP} is a widely employed privacy enhancement in cross-device FL~\cite{mcmahan2017learning}, and the FedAvg embedded with the \textit{client-level DP mechanism} is known as DPFed. Specifically, the local model update of the client $i$ in DPFed is clipped and perturbed by adding Gaussian noise drawn from the distribution $\mathcal{N}(0,{C^2\sigma^2})$ to each coordinate as follows:
\begin{equation}\label{eqn:modelperturbation}
     \Delta_i^t := \text{Clip}_C(\mathbf{x}^{t}-\mathbf{x}_i^{t,\tau} )  + \mathcal{N}(0,{C^2\sigma^2} \cdot \bm{I}_d),
\end{equation}
where $\sigma$ is the noise multiplier and $\text{Clip}_C(\mathbf{x}) = \mathbf{x} \times \min(1, {C}/{\|\mathbf{x}\|_2})$ represents a model clipping function with threshold $C$. The model perturbation described in \eqref{eqn:modelperturbation} is often used with \textit{secure aggregation}\cite{bonawitz2017practical} to amplify the client-level privacy guarantee in the cross-device FL setting. Secure aggregation ensures that the server only learns an aggregated function of the clients' local model updates, typically represented as the sum (i.e., $\sum_{i\in\mathcal{S}^t} \Delta_i^t$). To prevent privacy leakage from the sum of local model updates, Gaussian noise is added. By doing so, DPFed can achieve $(\epsilon, \delta)$-DP by choosing an appropriate noise multiplier $\sigma$ according to Theorem~\ref{thm:privacy_loss}.
\begin{theorem}[Privacy Guarantee of DPFed\cite{Fed-SMP}]
\label{thm:privacy_loss}
Suppose the client is sampled without replacement with probability $q:=s/n$ at each round. For any $\epsilon < 2\log(1/\delta)$ and $\delta\in(0,1)$, DPFed satisfies $(\epsilon, \delta)$-DP after $T$ communication rounds if
$\sigma^2 \geq {7q^2T(\epsilon + 2\log(1/\delta))}/{\epsilon^2}.
$
\end{theorem}

\section{Problems and Existing Solutions}\label{sec:sys-mod}
The vanilla FL training process (i.e., FedAvg described in Section~\ref{sec:fedvag}) is exposed to threats from both \textit{privacy attacker} and \textit{Byzantine attacker}. In this section, we define the {privacy attacker} and {Byzantine attacker} by specifying their attack goals, capabilities, and {existing solutions}.

\subsection{Problem Setting}
\textbf{The \textit{objective} of the privacy inference attacker} is to deduce sensitive information about clients' local data by observing the shared model updates. In terms of their \textit{capabilities}, these attackers could possess diverse prior information and computational resources, and they might collaborate with one another. However, their actions do not encompass injecting fabricated data into the system. These adversaries could be either clients or servers within the FL system, adhering to the designated training protocol but maintaining curiosity about the local data of others. To defend against such strong privacy attackers, the client-level DP mechanism is one of the mainstream \textit{solutions}. This mechanism leverages additive Gaussian noise to limit the amount of information that attackers can infer, providing a rigorous $(\epsilon, \delta)$-DP guarantee that ensures each client's privacy loss in FL is bounded by $\epsilon$ after training.

\textit{Given its rigorous privacy protection, we adopt DPFed~\cite{geyer2017differentially} - FedAvg embedded with client-level DP mechanism - as our base system for this study, with an aim to enhance its resilience against Byzantine attacks. More precisely, starting with the DPFed system that is guaranteed to achieve $(\epsilon, \delta)$-DP, our focus is on developing novel designs to improve its robustness against Byzantine attackers.}

\textbf{The \textit{objective} of a Byzantine attacker} within the FL is to undermine the overall performance of the trained global model as much as possible, also known as the {untargeted model poisoning attack}. In essence, the \textit{capability} of Byzantine clients involves maliciously altering their SGD processes and potentially collaborating with each other. However, they remain unaware of the model updates of benign clients due to the application of secure aggregation~\cite{kadhe2020fastsecagg}. In this study, we explore a more pragmatic situation wherein a Byzantine client might diverge from the standard local SGD process (as defined in Equation \eqref{eqn:local-sgd}), manipulating it to create harmful model updates. Notably, altering the local SGD process is a more feasible avenue for Byzantine attackers in DPFed than tampering with the DP mechanism, as the honesty verification method\cite{biswas2022verifiable} can be used to verify that the release DP statistic was computed correctly and the private randomness generated faithfully, even when using secure aggregation. Here, Byzantine clients can manipulate their local data or model parameters, causing the local SGD process to output malicious model updates. This potential threat emphasizes our investigation of devising Byzantine-robust strategies in the DPFed system, {ensuring a balance between data privacy and system integrity.}


\subsection{Byzantine Robustness of DPFed: Free but Limited!}

\begin{figure}
\begin{center}
\setlength{\fboxrule}{0pt}
\setlength{\fboxsep}{0cm}
\fbox{\includegraphics[width=\linewidth]{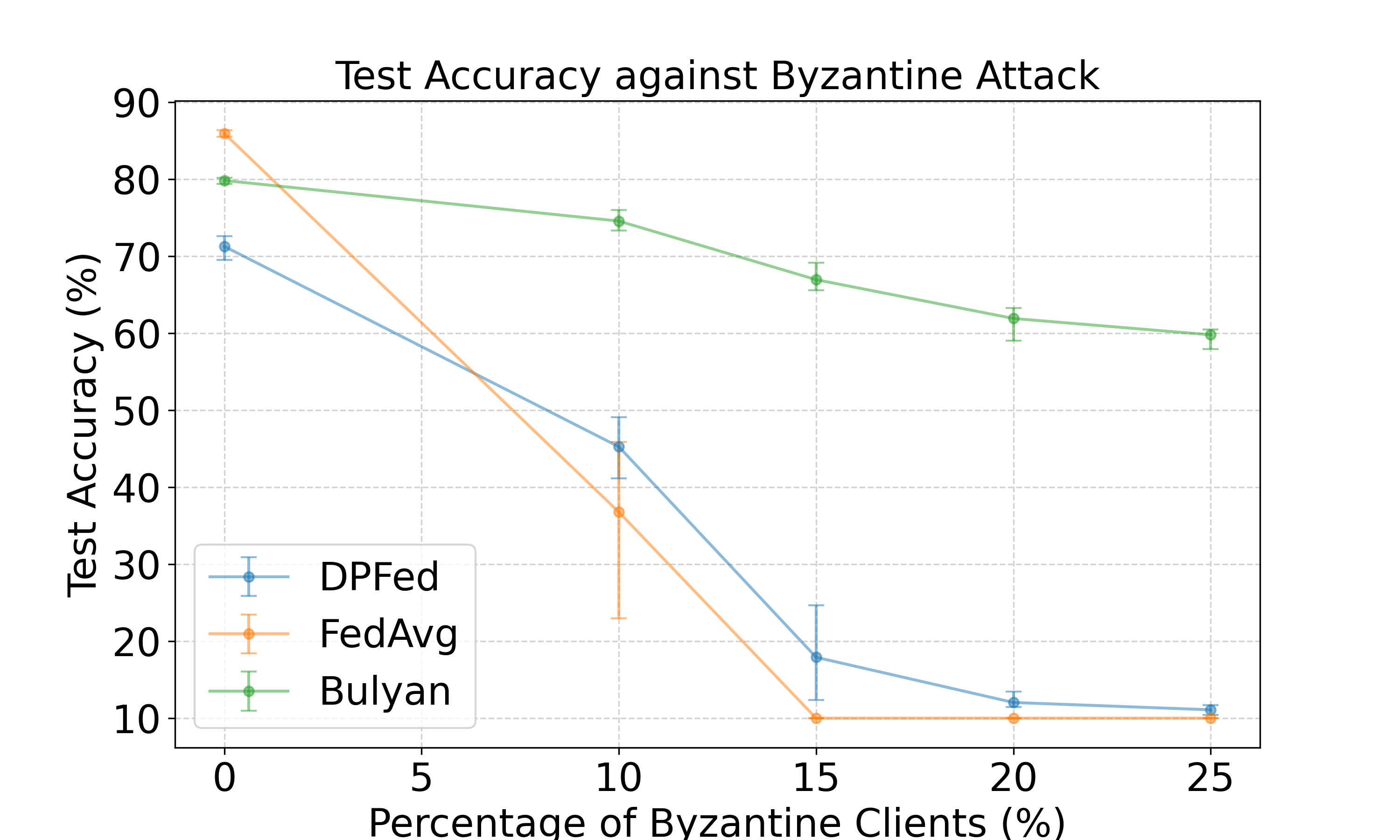}}
\end{center}
\vspace*{-5pt}
  \caption{Performances of FedAvg under Byzantine attack (AGR attack~\cite{agr} specifically) in the image classification task, compared with FedAvg and Bulyan~\cite{bulyan}. }
\label{fig:dp_ablation}
\vspace*{-19pt}
\end{figure}
Prior studies~\cite{can-you-really, flame} have demonstrated the effectiveness of incorporating ``weak" differential privacy to enhance model robustness through the introduction of noise during feature learning. Motivated by these findings, we investigate the potential benefits of leveraging higher DP noise levels to improve robustness. To evaluate this, We conduct experiments using the Fashion-MNIST~\cite{fmnist} and employ a 7-layer convolutional neural network (CNN). We specifically use AGR attack~\cite{agr}, a powerful Byzantine attack method that aims to divert the estimated benign aggregate in a malicious direction by generating a malicious perturbation. For further details, please refer to Section~\ref{sec:exp-settings}. In our evaluation, we test the client-level DP mechanism's robustness in DPFed against Byzantine attacks, comparing it with FedAvg and Bulyan~\cite{bulyan}. Bulyan is a Byzantine-robust aggregator that enhances the robustness of the FedAvg algorithm by identifying, screening, and treating local model updates that are deemed "benign" in the presence of Byzantine attacks. To ensure fairness in our comparison, we incorporate model clipping into the FedAvg baseline using the same threshold $C=1.0$. The key distinction between DPFed and the baseline method lies in the additive Gaussian noise. In our experiments, we set the noise multiplier $\sigma$ as 0.1 for DPFed. We report the average results over 5 runs.

Figure~\ref{fig:dp_ablation} illustrates that, at low Byzantine client percentages (PBC), FedAvg outperforms DPFed. This result indicates that while adding DP noise protects privacy, it also causes a noticeable rise in system error. As the PBC increases, the test accuracy of FedAvg significantly decreases, with a marked performance decline observed at a PBC of 10\%. From this point onwards, DPFed outperforms FedAvg up to a PBC of 25\%, beyond which both FedAvg and DPFed completely diverge. Although DPFed provides robustness against the Byzantine attack when the PBC is small, its performance decreases quickly as the PBC increases. Moreover, DPFed is unable to outperform the Byzantine-robust aggregator, with a notable performance gap between them. This informs us that \textbf{while the client-level DP mechanism provides a certain degree of Byzantine-robustness for free, this benefit is quite limited.} These findings lead us to an important question: \textit{"How can we enlarge the robustness benefit of client-level DP mechanism while preserving the same privacy guarantee?"} 

\section{Achieving Byzantine-Robustness with Variance-
Reduced Client-level DP Mechanism}\label{sec:our}
In the rest of this paper, we answer the above question by introducing new designs to the client-level DP mechanisms and proposing a new FL algorithm that can achieve high Byzantine robustness and strong privacy guarantee simultaneously. 
\subsection{Importance of Variance Reduction}
We first analyze the variance or perturbations in the aggregated local updates of DPFed with Byzantine clients. Assume that up to $B$ ($B< n/2$) clients in the system are Byzantine. Let $\mathcal{B}$ denote the set of Byzantine clients, and $\mathcal{H}$ denote the set of benign clients. For ease of illustration, we consider the case of $\tau=1$ and full participation so that the local gradient of each client will be calculated, processed through the client-level DP mechanism, and then sent to the server for aggregation. Let $\bm{g}_i^{t}$ denote the stochastic local gradient of benign client $i\in\mathcal{H}$ in round $t$, $\bm{b}_j^{t}$ denote the arbitrary update of Byzantine client $j\in\mathcal{B}$, $\nabla f_i(\mathbf{x}^t)$ be the true local gradient of client $i\in[n]$, $\nabla f(\mathbf{x}^t)$ be the true gradient of $f(\mathbf{x}^t)$. The discrepancy between the aggregated local updates $\Delta^t:= (1/n) \sum_{i\in[n]}\Delta_i^t$ and the true gradient is decomposed as
\begin{multline}\label{eqn:variance}
    \mathbb{E}\|\Delta^t - \nabla f(\mathbf{x}^t)\|^2 
     \leq \frac{4}{n-B} \sum_{i\in\mathcal{H}}\mathbb{E}\left\|\bm{g}_i^{t} - \nabla f_i(\mathbf{x}^t) \right\|^2 \\ + \frac{2}{B} \sum_{j\in\mathcal{B}}\mathbb{E}\left\|\bm{b}_j^{t} \right\|^2 
   + \frac{dC^2 \sigma^2}{s} + 4\kappa^2.
\end{multline}
Here, the expectation $\mathbb{E}[\cdot]$ is taken over the mini-batches $\xi_i^r,\forall i\in[n], r\in\{0,\dots,\tau-1\}$ at round $t$ and the DP noise, and we assume that the true local gradient is bounded, i.e., $ \nabla f_i(\mathbf{x})\leq \kappa^2, \forall i\in[n]$. The detailed proof is given in the appendix\footnote{https://goo.by/CxcGT}. From the error decomposition result in \eqref{eqn:variance}, we make the following observations: 
\begin{itemize}
    \item During Byzantine attacks, the aggregated model update error mainly arises from benign clients' local stochastic gradient variance (first term), Byzantine perturbation (second term), and DP perturbation (third term).
    \item DP perturbation can be significantly large for large models as it is proportional to the model dimension $d$.
    \item The use of model clipping in DPFed can restrict the malicious perturbations from Byzantine attackers. More specifically, with clipping, the malicious model can be bounded as $\mathbb{E}\left\|\bm{b}_j^{t} \right\|^2 \leq C$.
    \item Both DP and Byzantine perturbations can be mitigated by choosing a small clipping threshold. But, the variance of the gradient will increase if $C$ is too small to make the stochastic gradient deviate from the true local gradient. 
\end{itemize}
These insights shed light on how the interplay of different variances can influence the accuracy of the aggregated model update in DPFed under Byzantine attacks, which informs our new design for DPFed to improve Byzantine robustness. 

\subsection{Our Approach}
Inspired by these observations, we introduce our approach that employs sparsification and momentum techniques to decrease the variance of the aggregated update effectively.

\subsubsection{Local Variance Reduction with Momentum}
To improve the accuracy of the aggregated update, we first reduce the variance of stochastic gradients using local momentum. Momentum SGD and its variants, such as Adam~\cite{kingma2014adam}, have proven to be highly effective in various ML applications by reducing the variance of stochastic gradients. In light of these advantages, we adopt momentum SGD as the local optimizer on the client side to mitigate the total variance of the aggregated update and accelerate the convergence of our algorithm. 

In the context of cross-device FL, clients participate periodically, and there is a possibility that the local momentum histories from the current training round may become outdated for future training, especially when the client sampling rate is low. Thus, initializing local momentum at each training round's start becomes essential for its effectiveness.

\subsubsection{Mitigating DP and Byzantine Perturbations with Sparsification}
Next, we focus on mitigating the errors arising from DP and Byzantine perturbations. To reduce the DP perturbation in large deep neural network (DNN) models, we co-design top-$k$ sparsification with the client-level DP mechanism in DPFed. This integration is motivated by the dimension reduction property of model sparsification, as the DP perturbation is proportional to the model dimension $d$, which can be large for DNN models. In our approach, the client applies top-$k$ sparsification to its model update before the client-level DP mechanism, retaining only the top-$k$ parameters with the largest absolute values and setting the rest to zero, as defined in Definition~\ref{def:sparsifier}. During the DP mechanism, only the selected $k$ parameters are subjected to perturbation by adding Gaussian noise. Consequently, the DP perturbation error decreases from $dC^2\sigma^2/s$ to $kC^2\sigma^2/s$.

\begin{definition}[Top-$k$ Sparsification]\label{def:sparsifier}
For a parameter $1\leq k\leq d$ and vector $\mathbf{x}\in\mathbb{R}^d$, the top-$k$ sparsifier $\text{top}_k:\mathbb{R}^d \rightarrow \mathbb{R}^d$  are defined as: 
$
[\text{top}_k(\mathbf{x})]_j=
[\mathbf{x}]_{\pi(j)} \text{ if } j\leq k \text{, otherwise } [\text{top}_k(\mathbf{x})]_j = 0,
$
where $\pi$ is a permutation of $[d]$ such that $|[\mathbf{x}]_{\pi(j)}| \geq |[\mathbf{x}]_{\pi(j+1)}|$ for $j\in[1,d-1]$.
\end{definition}

Besides the DP error reduction, sparsification can also help to reduce the Byzantine perturbations. Intuitively, sparsification limits the number of parameters the attacker can alter, which reduces the attacker's action space. Theoretically, as $\mathbb{E}\left\|top_k(\bm{b}_j^{t}) \right\|^2 \leq \mathbb{E}\left\|\bm{b}_j^{t} \right\|^2$, using sparsification on the model updates allows us to use a smaller clipping threshold $C$ so that both Byzantine perturbation and DP perturbation are mitigated. 

Privacy is vital when combining top-$k$ sparsification and the client-level DP mechanism. Sorting private values during coordinate selection entails concealing private information within the chosen $k$ indexes. Moreover, it is also necessary to force Byzantine clients to follow the sparsification protocol. To address these challenges, we adopt a protocol where the selection of the top $k$ coordinates occurs globally on the server using the privacy-preserving global model. This design choice compels clients to consistently submit the same $k$ coordinates to the server in each training round without privacy violation. Note that Byzantine clients can arbitrarily manipulate and send $k$ coordinates to the server. Any Byzantine client that deviates from the sparsification protocol - for instance, by sending more than $k$ parameters - can be easily identified and rectified.

\setlength{\textfloatsep}{3pt}
\begin{algorithm}[htpb]
\caption{FedVRDP: Federated Learning with Variance-Reduced Client-level DP Mechanism}\label{algorithm-ours}
\begin{algorithmic}[1]
\REQUIRE parameters $n$, $s$, $\eta,\alpha$, $C$, $\sigma$, $\tau$, $T$, $k$, initial point $\mathbf{x}^0$
\STATE Generate initial mask $\bm{m}^{0}$ randomly
\FOR{$t=0$ to $T-1$}
    \STATE Server randomly samples a set of $s$ clients (denoted by $\mathcal{S}^t$) and broadcasts $\mathbf{x}^t$ and $\bm{m}^{t}$ to them
    \FOR{each clients $i \in \mathcal{S}^t$ \textbf{in parallel}}
        \STATE $\textbf{y}_i^t \leftarrow \textbf{VRDP-Update}(\mathbf{x}^t, \bm{m}^{t})$ 
    \ENDFOR
    \STATE $\mathbf{x}^{t+1} \leftarrow \mathbf{x}^t - (1/s)\sum_{i \in \mathcal{S}^t} {\textbf{y}_i^{t}} $
    \STATE $\bm{m}^{t+1} \leftarrow \text{TopMask}(\mathbf{x}^{t+1}, k) \hfill\lhd$ Generate mask
    \ENDFOR
    \returnx $\mathbf{x}^{T}$
\vspace*{4pt}
\VRDPClientUpdate
    \STATE $\mathbf{x}_i^{t, 0} \gets \mathbf{x}^t$, ${\bm{v}}_i^{t, 0} \gets \bm{0}_d $
    \FOR{$r=0$ to $\tau-1$}
        \STATE Compute a mini-batch stochastic gradient $\bm{g}_i^{t,r}$
        \STATE ${\bm{v}}_i^{t, r+1} \leftarrow (1-\kappa)\bm{v}_i^{t, r} - \kappa  \bm{g}_i^{t,r}$
        \STATE $\mathbf{x}_i^{t, r+1} \leftarrow \mathbf{x}_i^{t, r} - \eta  \bm{v}_i^{t, r} \hfill\lhd$ Momentum SGD
    \ENDFOR \label{alg1:ln_4}
    \STATE $\Delta^\prime \leftarrow (\mathbf{x}^t-\mathbf{x}_i^{t, \tau}) \odot \bm{m}^{t} \hfill\lhd$ Sparsify local update
    \STATE $\Delta_i^t \leftarrow \text{Clip}_C(\Delta^\prime)  + \mathcal{N}(0, {C^2\sigma^2}\cdot \bm{I}_k) \hfill\lhd $ Clip \& perturb
    \returnx $\Delta_i^t$
\end{algorithmic}
\end{algorithm}

\subsubsection{Our Algorithm}
Based on the aforementioned building blocks, we propose our algorithm, FedVRDP (\textbf{Fed}erated Learning with \textbf{V}ariance-\textbf{R}educed Client-level \textbf{DP} Mechanism), and outline the pseudo-code in Algorithm~\ref{algorithm-ours}. Specifically, in the $t$-th round of the training, the selected clients download the current global model $\mathbf{x}^t$ along with a binary sparsification mask $\mathbf{m}^{t}$. The mask vector $\mathbf{m}^t \in \{0,1\}^d$ is used to record the top $k$ coordinates for sparsification, and its $j$-th coordinate equals 1 if that coordinate is selected to be kept at round $t$ and 0 otherwise. The mask vector $\mathbf{m}^t$ is initialized by randomly selecting $k$ coordinates before the beginning of the first round of training. Subsequently, it is determined by calculating the top $k$ coordinates of the global model. Upon receiving the global model and mask, selected clients conduct the variance-reduced client-level DP update (i.e., VRDP-Update) in parallel on their local datasets. Specifically, each client performs momentum SGD for $\tau$ iterations (i.e., lines 12-16) to obtain a global model update $\mathbf{x}^t-\mathbf{x}_i^{t, \tau}$ locally. This local update is then sparsified using the mask (line 17). Here $\odot$ represents the dot product of two vectors, so only the values of the selected $k$ coordinates will be kept on the local update, and other values will be zeroed out. After sparsification, the sparsified model update is clipped using threshold $C$ with a commonly used clipping strategy~\cite{dp_original}, to mitigate the presence of abnormal parameters. To achieve client-level DP, the clipped model update will be perturbed by adding Gaussian noise (line 18). Note that only the non-zero elements in the clipped model update will be perturbed by DP noise and sent to the server via secure aggregation. For ease of expression, the secure aggregation protocol is not explicitly described in Algorithm~\ref{algorithm-ours} as it is not in our design scope. Still, we provide the details of the secure aggregation protocol~\cite{bonawitz2017practical} in the appendix. Finally, on the server side, local updates of selected clients are averaged and used to update the global model (line 7), and a new sparsification mask is generated for the next round by calculating the top $k$ coordinates of the updated global model, i.e., the TopMask step in line 8. 

Notably, the newly proposed local momentum update and sparsification process do not violate any privacy guarantee of the client-level DP mechanism. Indeed, our algorithm achieves exactly the same privacy guarantee as DPFed. By choosing $ \sigma \geq \sqrt{{7(s/n)^2T(\epsilon + 2\log(1/\delta))}/{\epsilon^2}} $, our algorithm achieves $(\epsilon,\delta)$ client-level DP for each client after $T$ rounds of training, according to Theorem~\ref{thm:privacy_loss}.

\section{EXPERIMENTAL EVALUATION}\label{sec:exp}
In this section, we assess the effectiveness of our method in mitigating Byzantine attacks and showcase its ability to achieve Byzantine-robustness and privacy guarantee concurrently, all while incurring no additional computing costs.

\subsection{Experimental Settings} \label{sec:exp-settings}

\begin{table*}[htpb]
\centering
\captionof{table}{Datasets, Models, and Learning Configurations.}
\resizebox{0.9\linewidth}{!}{
\begin{tabular}{c|ccccccc}
\hline
\textbf{Dataset}       & \textbf{Distribution} & \textbf{Client} & \textbf{Client/Round} & \textbf{Samples/Client} & \textbf{Model} & \textbf{Local Epoch} & \textbf{Global Round} \\ \hline
Fashion-MNIST & IID       & 6000   & 100      & 10                & CNN   & 10          & 180          \\
Shakespeare   & non-IID   & 715    & 100      & 52                & RNN   & 1           & 1000         \\ \hline
\end{tabular}
}
\label{table:exp-setting}
\vspace*{-5pt}
\end{table*}

\textbf{Datasets and Models.} We assess our approach using two common benchmark datasets (listed in Table~\ref{table:exp-setting}) for differentially private machine/federated learning. Fashion-MNIST~\cite{fmnist}, a widely employed image classification dataset, comprises 10 image categories with 60,000 training samples and 10,000 testing samples. Each grayscale image is 28 × 28 pixels. Our image classification task uses a CNN model with two 5 × 5 convolutional layers, followed by 2 × 2 max pooling and ReLU activation. It includes a fully-connected layer with a 512-dimensional output and a classifier head. This model has approximately $1.6$ million parameters. Our experiments employ an IID and cross-device setup, dividing the dataset into 6,000 clients, each with 10 training samples.

The Shakespeare dataset~\cite{fl_original, shakespeare} is a natural non-IID federated dataset for text generation tasks. The dataset consists of 37784 data samples from 715 clients, each representing a speaking role with at least two lines. We use a recurrent neural network (RNN) model for this dataset, which takes a sequence of characters as input and employs an embedding layer to transform each character into an 8-dimensional feature representation. These embedded characters are then processed through two LSTM layers, each consisting of 256 nodes. Finally, a densely connected softmax output layer is applied. We adopt a vocabulary size of 90 in our experiments. Our model is trained to predict a sequence of 80 characters by taking in a sequence of 80 characters, where the input sequence is shifted by one position. Consequently, the model has an output dimension of 80 × 90, totaling around 0.8 million parameters.

\textbf{Configurations.} Our experiments are conducted using PyTorch and executed on NVIDIA RTX A6000 GPUs. For both datasets, the server randomly selects $s=100$ clients to participate in training during each round. We employ the SGD with momentum as the local optimizer for our methods. Specifically, for Fashion-MNIST, we set the momentum coefficient as 0.5, and the local learning rate ($\eta$) is 0.125 and decays at a rate of 0.99 in each round. The batch size is set as 10, and the local epoch is set as 10, i.e., $\tau=10$. For Shakespeare, local momentum is set at 0.9 with a local learning rate of 1.0, decaying at 0.99 every 50 rounds. Batch size is 4, and local epochs are 1. We set $T=180$ and $T=1000$ for Fashion-MNIST and Shakespeare, respectively.

We calculate the end-to-end privacy loss using the API in \cite{opacus}. We set the privacy parameter $\delta$ following the methodologies employed in \cite{reddit} and \cite{andrew2021differentially}. The default value of the noise multiplier in our algorithm is set as $\sigma=1.4$ for both datasets. For the clipping threshold, as we mentioned, it can reduce the Byzantine and DP perturbation, but it will also lead to an increased gradient variance if it is too small. The model compression in our approach allows us to use a smaller clipping threshold. Therefore, for our approach, we tune the clipping threshold and model compression ratio (defined as $p:=k/d$) for both datasets by doing a grid search and finally have $p=0.3, C=0.5$ for the Fashion-MNIST dataset and $p=0.3, C=1.0$ for the Shakespeare dataset.

\textbf{Baselines.} We compare our approach with the state-of-the-art baselines, including four Byzantine-robust aggregators (namely Trimmed Mean~\cite{trmean_median}, Median\cite{trmean_median}, Krum~\cite{krum} and Bulyan~\cite{bulyan}), SparseFed~\cite{sparsefed}, which uses global model sparsification to achieve robustness against model poisoning attacks, and Flame \cite{flame} which also applies DP noise to improve robustness against model poisoning attacks. 
{The Byzantine-robust aggregators basically replace the averaging step of FedAvg with a robust aggregation rule.} For example, in the Trimmed Mean method, the server collects the values of a specific model parameter from all local model updates received from the clients and arranges them in ascending order. To enhance robustness, it removes extreme values using a parameter $f$. {In the experiments, we set the robustness parameter $f=10$.} {SparseFed} is a defense method based on sparsification. It employs top-$k$ sparsification on the aggregated model update and incorporates model clipping and global momentum with error feedback to defend against model poisoning attacks. {Flame} is close to our work in privacy protection, which utilizes model clustering and DP mechanism on the server side to defend against model poisoning attacks. It assumes that the server is trusted.

\textbf{Attacks.} 
We implement two Byzantine attacks, namely the Fang attack~\cite{fang} and the AGR attack~\cite{agr}. Fang attack is an aggregator-known attack method that requires the knowledge of the server's aggregator. On the other hand, the AGR attack provides a general framework for Byzantine attacks, making it applicable to optimize attacks for any given aggregation rule, regardless of whether full knowledge or partial knowledge is available. Both attack methods formulate the attack as an optimization problem, aiming to maximize the deviation of the global model update in the opposite direction of the benign updates. To ensure comprehensive evaluation, we consider the most powerful versions of the attacks based on the principles of the Cannikin Law (or Wooden Bucket Theory). For instance, we utilize the Fang attack specifically designed for Krum to evaluate the Krum and Bulyan defense methods, and we use the AGR attack designed specifically for Median to evaluate the performance of Median. In short, we always select the Fang/AGR attack with the best-attacking performance to evaluate the robustness of these defending methods. 

\subsection{Experimental Results}

\textbf{Effectiveness of FedVRDP.}
We first evaluate the effectiveness of our method in defending against the Fang attack and AGR attack, aiming to demonstrate its superiority over existing defense approaches. Table~\ref{table:exp-fmnist} presents the test accuracy results obtained when subjecting our method to a Byzantine attack on the Fashion-MNIST dataset. We consider a powerful attack setting where 20\% of the clients in the FL system are Byzantine. We also note the experiment's runtime (i.e., the duration) and report averaged results from 5 trials.

From the results, we can observe that among the defense baselines, only the Bulyan shows some resistance to the Fang attack, potentially benefiting from its multi-iteration model cleaning approach. However, Bulyan is limited to scenarios with less than 25\% compromised clients, and it will become ineffective beyond that threshold. Regarding the AGR attack, the Trimmed Mean, Median, and SparseFed methods exhibit behavior similar to that of a random classifier, with an accuracy of around 10\%. In comparison, our method outperforms Bulyan and Krum by +15.43\% and +10.81\%, respectively. In terms of running time, our method demonstrates an average increase of 6.6 and 5.4 minutes per round compared to the No Defense setting and SparseFed, respectively. These results highlight the advantages of our approach over Bulyan, which requires significantly longer times of 69.9 and 68.7 minutes.

On the Shakespeare dataset, as demonstrated in Table~\ref{table:exp-shakespeare}, our method exhibits a significant performance advantage over other Byzantine-robust approaches when subjected to Fang attack. When facing AGR attack, our method achieves comparable results to Bulyan, with only a slight difference of 0.1\% in test accuracy. It is worth noting, however, that the total training duration of Bulyan is approximately three times longer than our approach. Furthermore, our method prioritizes privacy preservation, whereas Bulyan does not offer any privacy guarantee. As a result, Bulyan may not be suitable if system efficiency and client privacy are important considerations.

\begin{table}[htpb]
\centering
\captionof{table}{Performance of FedVRDP against Byzantine attacks on Fashion-MNIST, compared to state-of-the-art defenses.}
\resizebox{0.9\linewidth}{!}{
\begin{tabular}{c|c|c|c}
\hline
\textbf{Defense}                            & \textbf{Fang(\%)}                      & \textbf{AGR(\%)}              &\textbf{Duration(Hrs)}\\ \hline
FedAvg                                      & \multicolumn{2}{c|}{85.94}                                              & 0.24              \\ \hline
No Defense                                  &  35.37                                 & 10.00                             & 0.25/0.25      \\ \hline
Trimmed Mean~\cite{trmean_median}            &  40.33                                & 11.05                             & 0.33/0.29         \\
Median~\cite{trmean_median}                        &  35.37                                 & 10.00                             & 0.31/0.26         \\
Krum~\cite{krum}                            &  29.44                                 & 66.53           & 0.28/0.46         \\
Bulyan~\cite{bulyan}                        &  63.17               & 61.91                             & 1.38/1.45         \\
SparseFed~\cite{sparsefed}                  &  23.76                                 & 10.00                             & 0.28/0.26         \\ \hline
\textbf{Ours}                               &  \textbf{74.62}                & \textbf{77.34}            & 0.41/0.31         \\ \hline
\end{tabular}
}
\label{table:exp-fmnist}
\vspace*{-5pt}
\end{table}

\begin{table}[htpb]
\vspace*{-10pt}
\centering
\captionof{table}{Performance of FedVRDP against Byzantine attacks on Shakespeare, compared to state-of-the-art defenses.}
\resizebox{0.9\linewidth}{!}{
\begin{tabular}{c|c|c|c}
\hline
\textbf{Defense}                            & \textbf{Fang(\%)}         & \textbf{AGR(\%)}                  & \textbf{Duration(Hrs)}       \\ \hline
FedAvg                                      & \multicolumn{2}{c|}{64.75}                                              & 2.43              \\ \hline
No Defense                                  & 34.77                 & 36.86                         &  2.43/2.43             \\ \hline
Trimmed Mean~\cite{trmean_median}            & 39.05                 & 46.16                         &  2.76/2.60          \\
Median~\cite{trmean_median}                        & 35.18                 & 37.18                         &  2.67/2.45          \\
Krum~\cite{krum}                            & 41.88                 & 48.18                         &  2.50/2.81             \\
Bulyan~\cite{bulyan}                        &44.48 & \textbf{55.94}       &  7.81/8.10         \\
SparseFed~\cite{sparsefed}                  & 40.25                 & 34.85                         &  3.12/2.60             \\ \hline
\textbf{Ours}                               &\textbf{48.12} & 55.84                              &  2.96/2.82             \\ \hline
\end{tabular}
}
\label{table:exp-shakespeare}
\vspace*{-5pt}
\end{table}

\begin{table}[b]
\vspace*{-5pt}
\caption{Privacy and accuracy performance of our approach under Byzantine attacks, compared with the related methods. Results marked with * indicate that the training was interrupted, and we do not report values of ${\epsilon}$ exceeding 50.}\label{table:exp-privacy}
\begin{center}
\resizebox{\linewidth}{!}{
\begin{tabular}{c|c|ccc|ccc}
\hline
                                   &                                                  & \multicolumn{3}{c|}{\textbf{F-MNIST}}                                                          & \multicolumn{3}{c}{\textbf{Shakespeare}}                                                       \\ \cline{3-8} 
\multirow{-2}{*}{\textbf{Defense}} & \multirow{-2}{*}{$\bm{\sigma}$} & $\bm{\epsilon}$ & \textbf{Fang(\%)}                   & \textbf{AGR(\%)}                   & $\bm{\epsilon}$ & \textbf{Fang(\%)}                   & \textbf{AGR(\%)}                   \\ \hline
                                   & 0.001                                            & -                                & \textbf{69.38} & 73.27 & -                                & 42.19 & 54.67                        \\
                                   & 0.01                                             & -                                & 19.58                        & 73.48                        & -                                & 41.63*                       & 54.84 \\
\multirow{-3}{*}{Flame~\cite{flame}}            & 0.10                                              & -                                & 8.44*                        & 7.51*                        & -                                & 1.36*                        & 1.39*                        \\ \hline
DPFed                              & 0.10                                              & 2.02                              & 11.37                        & 11.09                        & 11.51                             &29.27                              & 39.95                             \\ \hline
\textbf{Ours}                      & 0.14                                             & 1.01                              & 65.34 & \textbf{74.97} & 6.99                             & \textbf{42.38} & \textbf{54.90} \\ \cline{1-8}
\end{tabular}
}
\end{center}
\vspace*{-8pt}
\end{table}

\textbf{Privacy Guarantee.} In terms of privacy protection, we compare the test accuracy of our method with that of DPFed and Flame~\cite{flame}, both of which utilize DP techniques to ensure system privacy or robustness. In our method, we apply a noise multiplier $\sigma$ of 0.14 to achieve the desired privacy level. Similarly, for DPFed, a noise multiplier $\sigma$ of 0.10 is applied. In the case of Flame, we conduct experiments with various noise multipliers and calculate its privacy loss~\cite{flame}. Our experiment encompasses both the Fashion-MNIST and Shakespeare datasets, allowing us to assess the performance of our method under Byzantine attacks while operating within a limited privacy budget. Specifically, we apply the Fang attack and AGR attack with 25\% Byzantine clients to evaluate the robustness and privacy guarantees of our approach. 

As shown in Table~\ref{table:exp-privacy}, DPFed achieves a strong differential privacy guarantee, with privacy losses bounded by $\epsilon=2.02$ and $11.51$ for Fashion-MNIST and Shakespeare, respectively. However, it lacks resilience against Byzantine attacks when a significant number of clients exhibit malicious behavior, resulting in outputs that are nearly random. On the other hand, Flame demonstrates robustness against Byzantine attacks only when the level of differential privacy noise is trivial. It becomes vulnerable to compromise when the magnitude of the noise is substantial, leading to a failure to provide a satisfactory privacy guarantee. In contrast, our method excels in achieving excellent performance in both privacy protection and Byzantine robustness across both the image classification and word generation tasks.

\begin{figure}[t]
\centering
\includegraphics[width=0.5\textwidth]{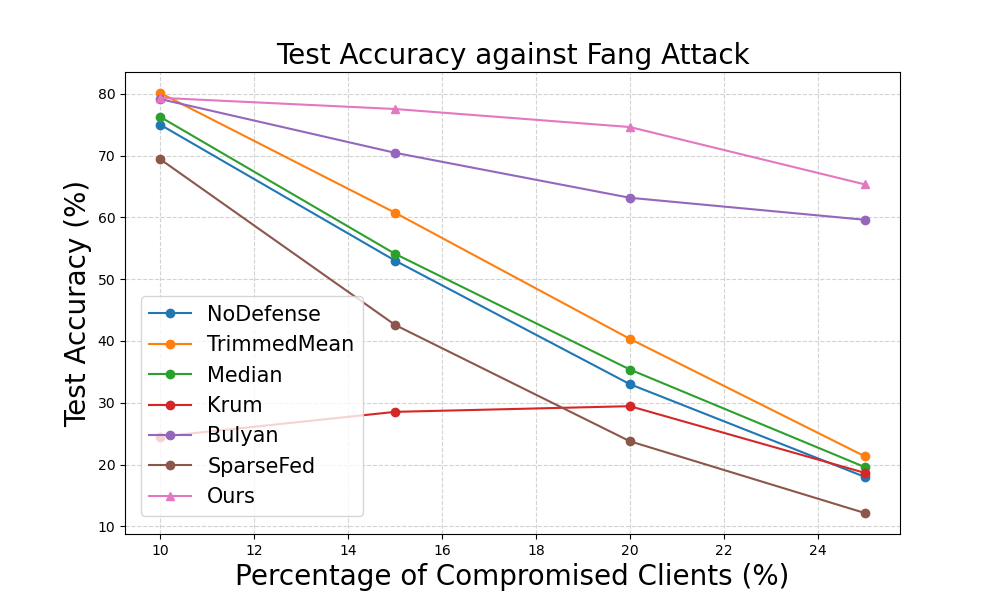}
\vspace*{-16.5pt}
\caption{Impact of the percentage of Byzantine clients on the performance of our approach against Fang attack, compared with the state-of-arts.}
\label{fig:exp-PBC-Fang}
\vspace*{-45pt}
\end{figure}

\begin{figure}[t]
\vspace*{-10pt}
\centering
\includegraphics[width=0.5\textwidth]{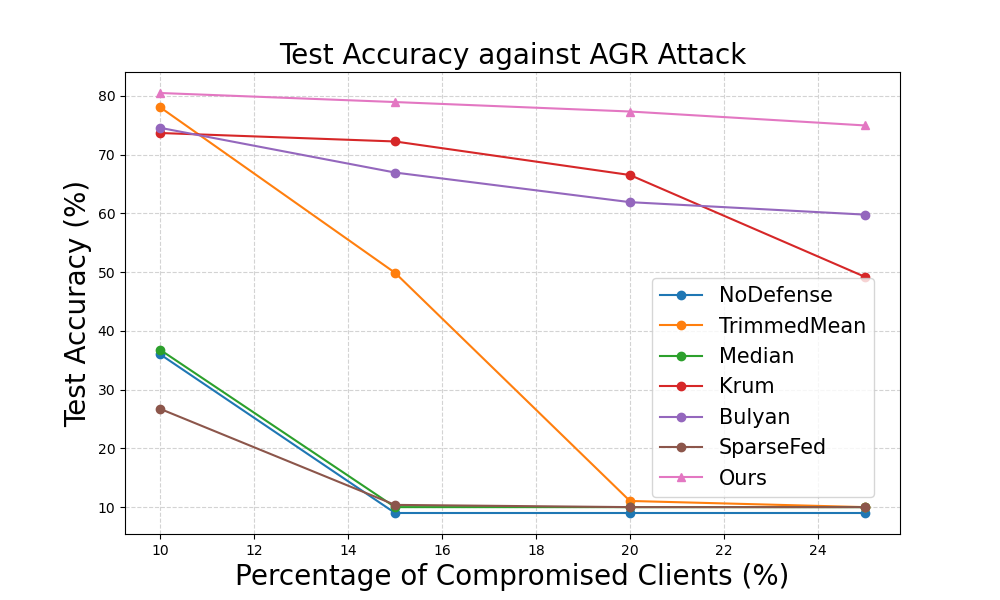}
\vspace*{-15pt}
\caption{Effect of Byzantine client percentage on our approach's performance against AGR attack, compared to STOA methods.}
\label{fig:exp-PBC-AGR}
\vspace*{-4pt}
\end{figure}

\textbf{Impact of Byzantine Clients Percentage.} We evaluate the performance of our method and baselines with respect to different percentages of Byzantine clients. Specifically, we vary the percentage of Byzantine clients from 10\% to 20\% and report the corresponding testing accuracy of the defense methods on Fashion-MNIST dataset in Figure~\ref{fig:exp-PBC-Fang} for the Fang attack and Figure~\ref{fig:exp-PBC-AGR} for the AGR attack, respectively. 
We observe that as the percentage of Byzantine clients increases from 10\% to 20\%, our method, along with Bulyan, can always maintain a test accuracy above 60\% against Fang attack, while other methods exhibit accuracies below 60\% when the percentage of Byzantine clients exceeds 15\%. Our method also demonstrates stability against AGR attack and outperforms other methods by a significant margin. {As the percentage of Byzantine clients increases from 10\% to 20\%, our testing accuracy under AGR attack only drops by 3.15\%}. 


\section{Related Work}\label{sec:related}

\subsection{Byzantine-robust Federated Learning}
The byzantine-robust aggregators have garnered considerable attention in recent years as an effective defense mechanism against various distributed attacks in federated learning systems. Several approaches have been developed based on heuristic-based statistical hypotheses, aiming to distinguish Byzantine-compromised models from others. The method proposed by \cite{trmean_median} proposes utilizing the mean values of each model coordinate on trimmed model candidates, demonstrating stable performance in the face of overwhelming Byzantine attacks. Another approach by \cite{geo_median} leverages the concept of geometric median in a novel manner, offering enhanced sample efficiency and applicability across a broader range of parameters. Furthermore, \cite{marginalmedian} demonstrates that the marginal median exhibits linear time complexity and does not require robustness parameter $f$. Distance-based methods have also been explored, including the approach introduced by \cite{krum}, which relies on the $l_2$ distance between each model and others to identify the closest neighbor as the updated model. Additionally, \cite{bulyan} enhances previous aggregators by combining statistical hypotheses and distance metrics to reduce estimation errors, albeit at the cost of increased computational requirements. Besides, \cite{sparsefed} employs top-$k$ sparsification on the aggregated model update and incorporates model clipping and global momentum with error feedback to defend against model poisoning attacks. None of the aforementioned methods provide a differential privacy guarantee, as their primary focus lies in defending against Byzantine attacks without considering privacy preservation. In contrast, our method operates without making any assumptions and achieves efficient performance by incorporating a variance-reduced client-level differential privacy mechanism.

\subsection{Federated Learning with Differential Privacy}
Federated learning offers promise in reducing the need for transmitting sensitive data to a central server. However, privacy risks are noted within the FL framework. Notably, recent studies have demonstrated that adversaries can reconstruct private information from local models using inference and inversion attacks~\cite{shokri2017membership}. To address this concern, one common approach is to introduce artificial noise, with differential privacy (DP) mechanisms being a prominent example~\cite{hu2020personalized, hu2021concentrated, hu2020privacy}. DP-based FL approaches~\cite{geyer2017differentially, smc_dp_fl},~\cite{hu2020federated} aim to strike a balance between privacy and convergence performance during the training process. However, our approach aims to enhance the robustness of the DP-based FL system against Byzantine attacks while ensuring a strong privacy guarantee.
 
\subsection{Co-design of Robustness and Privacy.}
Several studies~\cite{so2020byzantine, smc_dp_fl} have investigated the use of secure multi-party computation (MPC) combined with sophisticated filters to address the challenges posed by \textit{Byzantine attack} and \textit{privacy inference attack} at the same time. However, these methods typically assume the presence of a trusted central server, which may not be practical in certain scenarios. Other approaches~\cite{can-you-really, flame} have attempted to enhance robustness by introducing DP noise. Nevertheless, these methods have encountered difficulties in maintaining the desired level of privacy when using ``weak" DP noise. \cite{guerraoui2021differential} prove that the direct composition of Byzantine robustness and DP noise in the training of large models leads to practical infeasibility. Follow-up works~\cite{zhu2022bridging, ma2022differentially} attempt to address the issue by proposing novel model aggregation protocols. However, both of them incurs large cumulative privacy loss after the training. In our approach, we leverage sparsification to amplify the robustness benefit of DP noise without degrading the privacy guarantee, achieving high robustness and privacy simultaneously. 



\section{Conclusion}\label{sec:con}
This paper presents FedVRDP, a novel FL scheme that combines sparsification- and momentum-driven variance reduction techniques with the client-level DP mechanism. FedVRDP is designed to defend against both Byzantine attacks and privacy inference attacks. The scheme maintains the privacy guarantee of state-of-the-art FL with client-level DP scheme while enhancing robustness against Byzantine clients. Thus, FedVRDP achieves Byzantine-robustness and rigorous privacy protection simultaneously. The numerical experiments demonstrate its successful defense against multiple Byzantine attacks and high privacy guarantees in both IID and non-IID datasets.



{
\small
\bibliographystyle{plain}
\bibliography{IEEEexample,hu}

\begin{thebibliography}{10}

\bibitem{andrew2021differentially}
Galen Andrew, Om~Thakkar, Brendan McMahan, and Swaroop Ramaswamy.
\newblock Differentially private learning with adaptive clipping.
\newblock {\em Advances in Neural Information Processing Systems}, 34:17455--17466, 2021.

\bibitem{biswas2022verifiable}
Ari Biswas and Graham Cormode.
\newblock Verifiable differential privacy for when the curious become dishonest.
\newblock {\em arXiv preprint arXiv:2208.09011}, 2022.

\bibitem{krum}
Peva Blanchard, El~Mahdi El~Mhamdi, Rachid Guerraoui, and Julien Stainer.
\newblock Machine learning with adversaries: Byzantine tolerant gradient descent.
\newblock {\em Advances in neural information processing systems}, 30, 2017.

\bibitem{bonawitz2017practical}
Keith Bonawitz, Vladimir Ivanov, Ben Kreuter, Antonio Marcedone, H~Brendan McMahan, Sarvar Patel, Daniel Ramage, Aaron Segal, and Karn Seth.
\newblock Practical secure aggregation for privacy-preserving machine learning.
\newblock In {\em proceedings of the 2017 ACM SIGSAC Conference on Computer and Communications Security}, pages 1175--1191, 2017.

\bibitem{geo_median}
Yudong Chen, Lili Su, and Jiaming Xu.
\newblock Distributed statistical machine learning in adversarial settings: Byzantine gradient descent.
\newblock {\em Proceedings of the ACM on Measurement and Analysis of Computing Systems}, 2017.

\bibitem{dp_original}
Cynthia Dwork and Aaron Roth.
\newblock The algorithmic foundations of differential privacy.
\newblock {\em Found. Trends Theor. Comput. Sci.}, 2014.

\bibitem{fang}
Minghong Fang, Xiaoyu Cao, Jinyuan Jia, and Neil~Z. Gong.
\newblock Local model poisoning attacks to byzantine-robust federated learning.
\newblock In {\em Proceedings of the 29th USENIX Conference on Security Symposium}, 2020.

\bibitem{geyer2017differentially}
Robin~C Geyer, Tassilo Klein, and Moin Nabi.
\newblock Differentially private federated learning: A client level perspective.
\newblock {\em arXiv preprint arXiv:1712.07557}, 2018.

\bibitem{guerraoui2021differential}
Rachid Guerraoui, Nirupam Gupta, Rafa{\"e}l Pinot, S{\'e}bastien Rouault, and John Stephan.
\newblock Differential privacy and byzantine resilience in sgd: Do they add up?
\newblock In {\em Proceedings of the 2021 ACM Symposium on Principles of Distributed Computing}, pages 391--401, 2021.

\bibitem{bulyan}
Rachid Guerraoui, S{\'e}bastien Rouault, et~al.
\newblock The hidden vulnerability of distributed learning in byzantium.
\newblock In {\em International Conference on Machine Learning}, pages 3521--3530. PMLR, 2018.

\bibitem{hu2020federated}
Rui Hu, Y.~Gong, and Y.~Guo.
\newblock Federated learning with sparsification-amplified privacy and adaptive optimization.
\newblock {\em arXiv:2008.01558}, 2020.

\bibitem{hu2020trading}
Rui Hu and Yanmin Gong.
\newblock Trading data for learning: Incentive mechanism for on-device federated learning.
\newblock In {\em GLOBECOM 2020-2020 IEEE Global Communications Conference}, pages 1--6. IEEE, 2020.

\bibitem{Fed-SMP}
Rui Hu, Yanmin Gong, and Yuanxiong Guo.
\newblock Federated learning with sparsified model perturbation: Improving accuracy under client-level differential privacy.
\newblock {\em CoRR}, abs/2202.07178, 2022.

\bibitem{hu2022energy}
Rui Hu, Y.~Guo, and Y.~Gong.
\newblock Energy-efficient distributed machine learning at wireless edge with device-to-device communication.
\newblock In {\em ICC 2022-IEEE International Conference on Communications}. IEEE, 2022.

\bibitem{hu2021concentrated}
Rui Hu, Yuanxiong Guo, and Yanmin Gong.
\newblock Concentrated differentially private federated learning with performance analysis.
\newblock {\em IEEE Open Journal of the Computer Society}, 2:276--289, 2021.

\bibitem{hu2020personalized}
Rui Hu, Yuanxiong Guo, Hongning Li, Qingqi Pei, and Yanmin Gong.
\newblock Personalized federated learning with differential privacy.
\newblock {\em IEEE Internet of Things Journal}, 2020.

\bibitem{hu2020privacy}
Rui Hu, Yuanxiong Guo, Hongning Li, Qingqi Pei, and Yanmin Gong.
\newblock Privacy-preserving personalized federated learning.
\newblock In {\em ICC 2020-2020 IEEE International Conference on Communications (ICC)}. IEEE, 2020.

\bibitem{hu2019targeted}
Rui Hu, Yuanxiong Guo, Miao Pan, and Yanmin Gong.
\newblock Targeted poisoning attacks on social recommender systems.
\newblock In {\em 2019 IEEE Global Communications Conference (GLOBECOM)}, pages 1--6. IEEE, 2019.

\bibitem{kadhe2020fastsecagg}
Swanand Kadhe, Nived Rajaraman, O~Ozan Koyluoglu, and Kannan Ramchandran.
\newblock Fastsecagg: Scalable secure aggregation for privacy-preserving federated learning.
\newblock {\em arXiv preprint arXiv:2009.11248}, 2020.

\bibitem{kingma2014adam}
Diederik~P Kingma and Jimmy Ba.
\newblock Adam: A method for stochastic optimization.
\newblock {\em arXiv preprint arXiv:1412.6980}, 2014.

\bibitem{advl}
Han Liu, Zhiyuan Yu, Mingming Zha, XiaoFeng Wang, William Yeoh, Yevgeniy Vorobeychik, and Ning Zhang.
\newblock When evil calls: Targeted adversarial voice over ip network.
\newblock In {\em Proceedings of the 2022 ACM SIGSAC Conference on Computer and Communications Security}, 2022.

\bibitem{ma2022differentially}
Xu~Ma, Xiaoqian Sun, Yuduo Wu, Zheli Liu, Xiaofeng Chen, and Changyu Dong.
\newblock Differentially private byzantine-robust federated learning.
\newblock {\em IEEE Transactions on Parallel and Distributed Systems}, 2022.

\bibitem{fl_original}
Brendan McMahan, Eider Moore, Daniel Ramage, Seth Hampson, and Blaise~Aguera y~Arcas.
\newblock Communication-efficient learning of deep networks from decentralized data.
\newblock In {\em Artificial intelligence and statistics}, pages 1273--1282. PMLR, 2017.

\bibitem{mcmahan2017learning}
H.~Brendan McMahan, Daniel Ramage, Kunal Talwar, and Li~Zhang.
\newblock Learning differentially private recurrent language models.
\newblock In {\em International Conference on Learning Representations}, 2018.

\bibitem{nasr2019comprehensive}
Milad Nasr, Reza Shokri, and Amir Houmansadr.
\newblock Comprehensive privacy analysis of deep learning: Passive and active white-box inference attacks against centralized and federated learning.
\newblock In {\em 2019 IEEE symposium on security and privacy (SP)}, pages 739--753. IEEE, 2019.

\bibitem{flame}
Thien~Duc Nguyen, Phillip Rieger, Roberta De~Viti, Huili Chen, Bj{\"o}rn~B Brandenburg, Hossein Yalame, Helen M{\"o}llering, Hossein Fereidooni, Samuel Marchal, Markus Miettinen, et~al.
\newblock $\{$FLAME$\}$: Taming backdoors in federated learning.
\newblock In {\em 31st USENIX Security Symposium (USENIX Security 22)}, pages 1415--1432, 2022.

\bibitem{sparsefed}
Ashwinee Panda, Saeed Mahloujifar, Arjun~Nitin Bhagoji, Supriyo Chakraborty, and Prateek Mittal.
\newblock Sparsefed: Mitigating model poisoning attacks in federated learning with sparsification.
\newblock In {\em International Conference on Artificial Intelligence and Statistics}. PMLR, 2022.

\bibitem{reddit}
H.~Brendan McMahan~Daniel Ramage, Kunal Talwar, and Li~Zhang.
\newblock Learning differentially private recurrent language models.
\newblock In {\em 6th International Conference on Learning Representations, {ICLR} 2018, Vancouver, BC, Canada, April 30 - May 3, 2018, Conference Track Proceedings}.

\bibitem{shakespeare}
Sashank~J. Reddi, Zachary Charles, Manzil Zaheer, Zachary Garrett, Keith Rush, Jakub Kone{\v{c}}n{\'y}, Sanjiv Kumar, and Hugh~Brendan McMahan.
\newblock Adaptive federated optimization.
\newblock In {\em 9th International Conference on Learning Representations, {ICLR} 2021}. OpenReview.net, 2021.

\bibitem{agr}
Virat S. and Amir H.
\newblock Manipulating the byzantine: Optimizing model poisoning attacks and defenses for federated learning.
\newblock In {\em NDSS}, 2021.

\bibitem{shokri2017membership}
Reza Shokri, Marco Stronati, Congzheng Song, and Vitaly Shmatikov.
\newblock Membership inference attacks against machine learning models.
\newblock In {\em IEEE Symposium on Security and Privacy (SP)}, pages 3--18. IEEE, 2017.

\bibitem{so2020byzantine}
Jinhyun So, Ba{\c{s}}ak G{\"u}ler, and A~Salman Avestimehr.
\newblock Byzantine-resilient secure federated learning.
\newblock {\em IEEE Journal on Selected Areas in Communications}, 39(7):2168--2181, 2020.

\bibitem{can-you-really}
Ziteng Sun, P.~Kairouz, A.~Theertha Suresh, and H.~B. McMahan.
\newblock Can you really backdoor federated learning?
\newblock {\em CoRR}, abs/1911.07963, 2019.

\bibitem{smc_dp_fl}
Stacey Truex, Nathalie Baracaldo, Ali Anwar, Thomas Steinke, Heiko Ludwig, Rui Zhang, and Yi~Zhou.
\newblock A hybrid approach to privacy-preserving federated learning.
\newblock In {\em Proceedings of the 12th {ACM} Workshop on Artificial Intelligence and Security, 2019}. {ACM}, 2019.

\bibitem{iotw}
Fei Wen, Mian Qin, P.~V Gratz, and AL~N. Reddy.
\newblock Hardware memory management for future mobile hybrid memory systems.
\newblock {\em IEEE Transactions on computer-aided design of integrated circuits and systems}, 2020.

\bibitem{fmnist}
Han Xiao, Kashif Rasul, and Roland Vollgraf.
\newblock Fashion-mnist: a novel image dataset for benchmarking machine learning algorithms.
\newblock {\em arXiv preprint arXiv:1708.07747}, 2017.

\bibitem{marginalmedian}
Cong Xie, Oluwasanmi Koyejo, and Indranil Gupta.
\newblock Generalized byzantine-tolerant sgd.
\newblock {\em arXiv preprint arXiv:1802.10116}, 2018.

\bibitem{trmean_median}
Dong Yin, Yudong Chen, R.~Kannan, and P.~Bartlett.
\newblock Byzantine-robust distributed learning: Towards optimal statistical rates.
\newblock In {\em International Conference on Machine Learning}. PMLR, 2018.

\bibitem{opacus}
Ashkan Yousefpour, Igor Shilov, Alexandre Sablayrolles, Davide Testuggine, Karthik Prasad, Mani Malek, John Nguyen, Sayan Ghosh, Akash Bharadwaj, Jessica Zhao, et~al.
\newblock Opacus: User-friendly differential privacy library in pytorch.
\newblock {\em arXiv preprint arXiv:2109.12298}, 2021.

\bibitem{commz}
Yijing Zeng.
\newblock {\em Towards Large-Scale Spectrum Sensing and Data Analysis}.
\newblock The University of Wisconsin-Madison, 2022.

\bibitem{zhu2022bridging}
Heng Zhu and Qing Ling.
\newblock Bridging differential privacy and byzantine-robustness via model aggregation.
\newblock {\em arXiv preprint arXiv:2205.00107}, 2022.

\end{thebibliography}
}

\end{document}